\documentclass[letterpaper, 10 pt, conference]{ieeeconf}  

\IEEEoverridecommandlockouts                              

\usepackage{graphics} 
\usepackage{epsfig} 
\usepackage{times} 
\usepackage{amsmath} 
\usepackage{amssymb}  
\usepackage{booktabs} 
\usepackage{url}
\usepackage{tabularx}
\usepackage{graphicx}
\usepackage{cuted}
\usepackage{capt-of}

\usepackage[hidelinks]{hyperref}
\usepackage{algorithm}
\usepackage{algpseudocode} 
\usepackage{float}
\usepackage{cite}
\usepackage{amsfonts}
\usepackage{xcolor}
\usepackage{soul}
\usepackage[switch]{lineno} 
\usepackage{url}
\usepackage{mathtools} 
\usepackage[capitalise]{cleveref}
\usepackage{romannum}

\xdefinecolor{redZ}{RGB}{234,29,93}
\xdefinecolor{blueZ}{RGB}{19,106,213}
\xdefinecolor{yellowZ}{RGB}{251,138,46}
\xdefinecolor{greenYL}{RGB}{0, 191, 0}

\definecolor{ResponseRed}{RGB}{200,0,0}

\title{\LARGE \bf
Safe-Night VLA: Seeing the Unseen via Thermal-Perceptive Vision-Language-Action Models for Safety-Critical Manipulation
}

\author{Dian~Yu$^{\ast}$, Qingchuan~Zhou$^{\ast}$, Bingkun~Huang, Majid~Khadiv, Zewen~Yang$^{\dagger}$%
\thanks{D. Yu, Q. Zhou, B. Huang, M. Khadiv, and Z. Yang are with Munich Institute of Robotics and Machine Intelligence (MIRMI), Technical University of Munich (TUM), 80992 Munich, Germany.}%
\thanks{$^{\ast}$Equal contribution. $^{\dagger}$Corresponding author. \textless zewen.yang@tum.de \textgreater
}%
}

\begin{document}

\maketitle
\thispagestyle{empty}
\pagestyle{empty}

\begin{abstract}
Current Vision-Language-Action (VLA) models rely primarily on RGB perception, preventing them from capturing modalities such as thermal signals that are imperceptible to conventional visual sensors. 
Moreover, end-to-end generative policies lack explicit safety constraints, making them fragile when encountering obstacles. 
To address these limitations, we propose Safe-Night VLA, a multimodal manipulation framework that enables robots to see the unseen while enforcing geometric constraints for thermal-aware manipulation in unstructured environments. 
Specifically, Safe-Night VLA integrates long-wave infrared thermal perception into a pre-trained vision-language backbone, enabling semantic reasoning guided by thermodynamic properties. 
To enforce geometric constraints during execution, we incorporate a safety filter via control barrier functions, which provide deterministic workspace constraint enforcement during policy execution. 
We validate our framework through real-world experiments on a Franka manipulator, introducing a novel evaluation paradigm featuring temperature-conditioned manipulation, subsurface target localization, and reflection disambiguation, while maintaining constrained execution at inference time. 
Results demonstrate that Safe-Night VLA outperforms RGB-only baselines and provide empirical evidence that foundation models can effectively leverage non-visible physical modalities for the safety-critical manipulation tasks. 
Project page is available at
\href{https://thisanwerss.github.io/Safe-Night-VLA}
{https://thisanwerss.github.io/Safe-Night-VLA}.

\end{abstract}

\section{INTRODUCTION}
State-of-the-art (SOTA) Vision-Language-Action (VLA) models have demonstrated impressive capability in grounding natural language into motor control within structured settings \cite{brohan2023rt2visionlanguageactionmodelstransfer,octomodelteam2024octoopensourcegeneralistrobot,nvidia2025gr00tn1openfoundation, intelligence2026pi07steerablegeneralistrobotic}. 
As robot manipulation transitions toward unstructured real-world environments, however, policies increasingly face the dual challenge of the ``unseen''. 
First, standard RGB sensors lack direct observability of intrinsic physical properties like surface temperature or sub-surface states, limiting the robot's ability to perform thermodynamic reasoning. 
Second, VLA models execute unpredictable actions when encountering out-of-distribution (OOD) scenes or workspace boundaries. 
Deploying reliable VLA systems therefore requires expanding the perceptual modality to capture hidden physical states, alongside strictly enforcing geometric safety during execution.

\begin{figure}[t]
    \centering
    \includegraphics[width=0.48\textwidth]{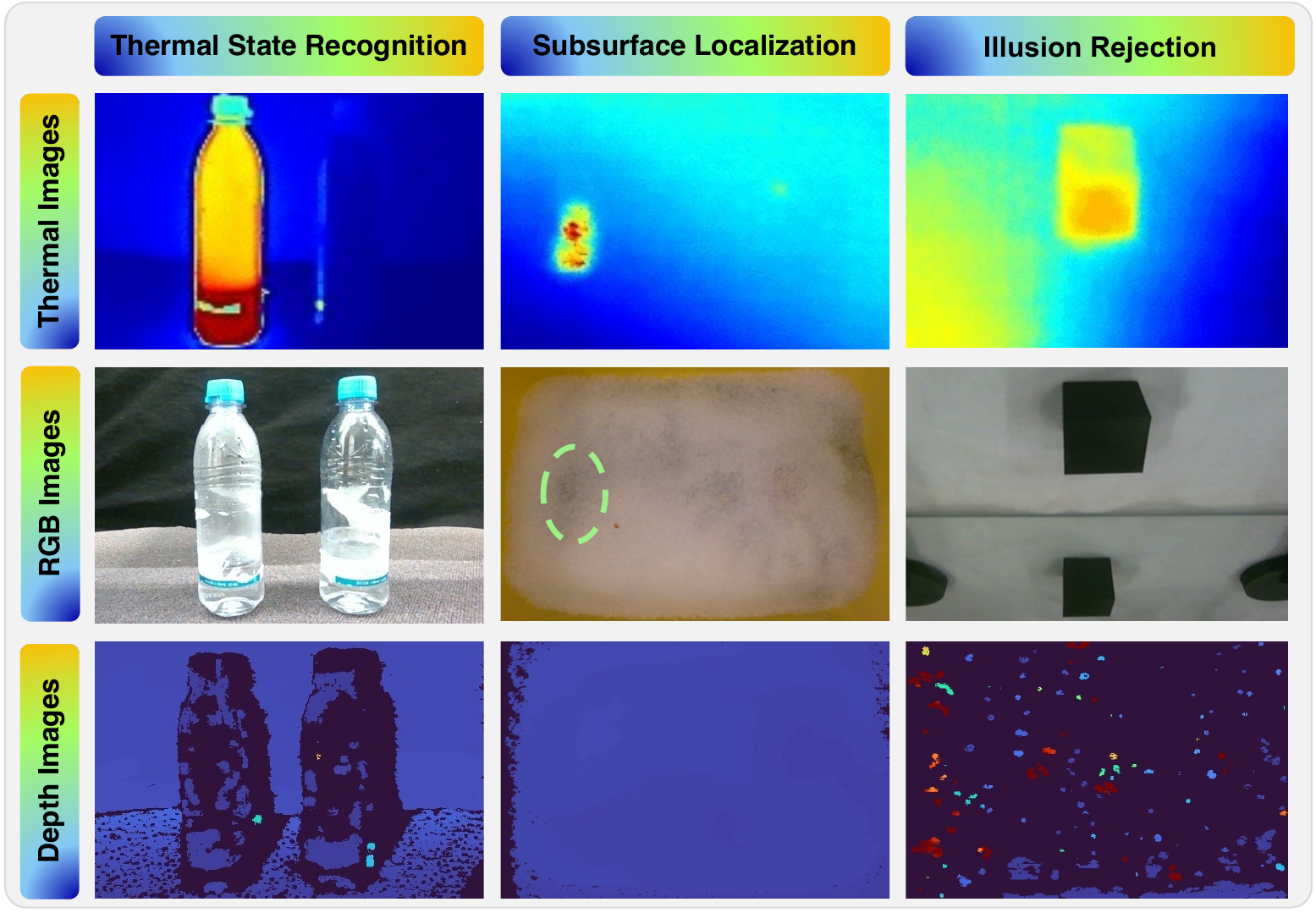} 
\caption{
Multimodal perception comparison in downstream tasks. LWIR thermal observations (top row) are shown alongside RGB (middle row) and depth images (bottom row). Temperature-aware recognition: distinguishing thermally distinct yet visually indistinguishable objects for handling of hot vs. cold items; 
Subsurface localization: detecting targets occluded beneath granular media; 
Illusion rejection: suppressing mirror-reflection artifacts by leveraging the LWIR attenuation of common glass.}
\label{fig:concept_teaser}
\end{figure}

To overcome these perceptual limitations, the community has begun moving beyond RGB sensing by incorporating depth \cite{zhen20243dvla3dvisionlanguageactiongenerative}, tactile feedback \cite{touch_vla}, and audio \cite{audio_robot}. 
Yet, Long-Wave Infrared (LWIR) thermal perception remains largely unexplored in VLA models. 
Unlike depth or tactile sensing, which primarily characterize geometry and contact, thermal imaging captures intrinsic physical properties that are fundamentally unobservable in the visible spectrum. 
As illustrated in Fig. \ref{fig:concept_teaser}, we formulate three representative scenarios where RGB policies are highly susceptible to visual aliasing or physical occlusions, whereas thermal perception successfully grounds the unseen states. 

However, expanded perception alone does not inherently prevent a neural policy from executing unsafe actions. 
A major limitation of current generative VLA policies is that they lack runtime safety guarantees and are vulnerable to unpredictable hallucinations, especially when facing optical artifacts or dynamic environments. 
Therefore, high-level semantic intent must be strictly decoupled from low-level geometric safety. 
To bridge this gap, we integrate control barrier functions (CBFs) \cite{ames2017control} as a runtime safety layer into the VLA framework. 
By bounding the VLA's output space at the control level, the CBF acts as a runtime geometric safeguard, intercepting policy hallucinations in OOD scenarios before they result in collisions.

To address these perceptual and safety limitations, we propose {Safe-Night VLA}, an empirical multimodal manipulation framework considering safe operations. 
The main contributions of this paper are threefold: 
\begin{itemize}
    \item We introduce the Safe-Night VLA framework, which integrates LWIR thermal perception into a frozen VLM and couples it with a CBF safety filter. It presents a pipeline that grounds natural language in thermodynamic properties while enforcing deterministic geometric constraints, addressing both the perceptual blind spots and execution vulnerabilities of standard policies. 
    \item We design a novel benchmark for three fundamental RGB failure modes: hidden thermodynamic states, subsurface target occlusion, and cross-modal optical deception, which provides a controlled environment for evaluating the robustness of multimodal VLA systems.
    \item Through cross-modal attention ablation studies, we reveal that the policy actively grounds semantic tokens in thermal gradients rather than relying on dataset-induced spatial biases, which offers new insights into how pre-trained vision models transfer to pseudo-color thermal domains without catastrophic forgetting.
\end{itemize}

\section{RELATED WORK}
The paradigm of modern robotic learning has shifted toward large-scale VLA models. Approaches such as RT-2, Octo, $\pi$, and GR00T~\cite{nvidia2025gr00tn1openfoundation, brohan2023rt2visionlanguageactionmodelstransfer, octomodelteam2024octoopensourcegeneralistrobot} demonstrate that fine-tuning Vision-Language Models (VLMs) on robotic action data yields strong semantic understanding and cross-task generalization. Yet, these models are predominantly built upon RGB observations, exposing them to the perceptual variability inherent in real-world deployments. 
Comprehensive evaluations~\cite{dey2025revlarevertingvisualdomain} systematically characterize the fragility of such policies under adverse lighting, visual clutter, and partial occlusions. 
Despite these findings, thermal infrared as a complementary sensing modality has been largely overlooked.  
Existing efforts narrowly focused on specific tasks such as thermal-aware grasping or outdoor navigation~\cite{hirose2025omnivla}, rather than addressing the general robustness of manipulation policies. 
Beyond photometric robustness, RGB observations are fundamentally limited by their inability to capture physically meaningful object states, such as surface temperature or material properties, that are critical for safe
physical interaction~\cite{guo2025omnivlaphysicallygroundedmultimodalvla}.

Thermal imaging has been applied to robotics primarily for low-level perception. 
Multimodal pipelines fusing RGB, depth, and thermal data have demonstrated improved grasp detection under low-light conditions~\cite{song2023vdt}, while thermal cues have been exploited to segment transparent objects, such as glass containers, that are effectively invisible to standard RGB sensors~\cite{transparent_review}. 
However, these approaches treat thermal data as a geometric or photometric aid, without coupling it to high-level semantic reasoning.
While recent concurrent work has begun integrating non-visual modalities such as depth and tactile sensing into foundation
models, thermal infrared remains absent from this line of research. Critically, existing thermal robotics methods lack the semantic grounding necessary to interpret natural language instructions, e.g., reasoning over an object described as ``the hot liquid'', requires jointly processing thermal state and linguistic context, a capability that current approaches do not support.

Bridging this gap requires reconciling the modality mismatch between thermal infrared and RGB-pretrained foundation models without sacrificing their pre-trained representations. Research in RGB-Thermal tracking shows that freezing a pre-trained RGB model and learning only lightweight thermal prompts achieves state-of-the-art performance with a fraction of the trainable parameters~\cite{vipt}. Complementary evidence from RGB-Thermal object detection further demonstrates that frozen RGB backbones can outperform fully fine-tuned counterparts, as freezing preserves the rich structural features encoded during pretraining~\cite{lu2026enhancing}.
This behavior is rooted in the shape bias of vision foundation models such as SigLIP~\cite{shapeclipper}, whose geometry-driven representations transfer effectively across spectral modalities. 
These findings motivate a parameter-efficient integration strategy: thermal semantics can be aligned into an existing VLA backbone through lightweight adaptation, preserving pre-trained world knowledge while acquiring temperature-aware perception.

As robot policies increasingly operate on physical hardware, preventing unsafe collisions arising from policy hallucinations becomes critical.
While early approaches addressed safety through training-time alignment via RLHF~\cite{autort2024}, recent methods have shifted toward enforcing rigorous geometric constraints at inference time. CBFs~\cite{ames2017control} have become a standard tool for guaranteeing forward-invariant safe behavior by restricting system states to a designated safe set.
Integrating CBFs with probabilistic generative policies introduces unique challenges. SafeDiffuser~\cite{xiao2025safediffuser} pioneered the embedding of CBF constraints directly into the diffusion denoising process to ensure collision-free planning.
More recently, hybrid architectures employing a VLM as a high-level safety monitor have been proposed~\cite{zhang2025safevla}, yet these methods typically suffer from high latency and offer no guarantees on low-level kinematic safety~\cite{dai_arxiv_2025_safeflowsaferobotmotion,yang_arxiv_2026_uniconflowunifiedconstrainedflowmatching}. Our approach adopts a CBF layer applied as a strictly post-hoc runtime filter on the DiT action head, which intervenes after action sampling and before hardware execution. 


\section{METHODOLOGY}

\subsection{System Architecture and Adaptation Strategy}
\begin{figure}[t]
    \centering
    \includegraphics[width=0.48\textwidth]{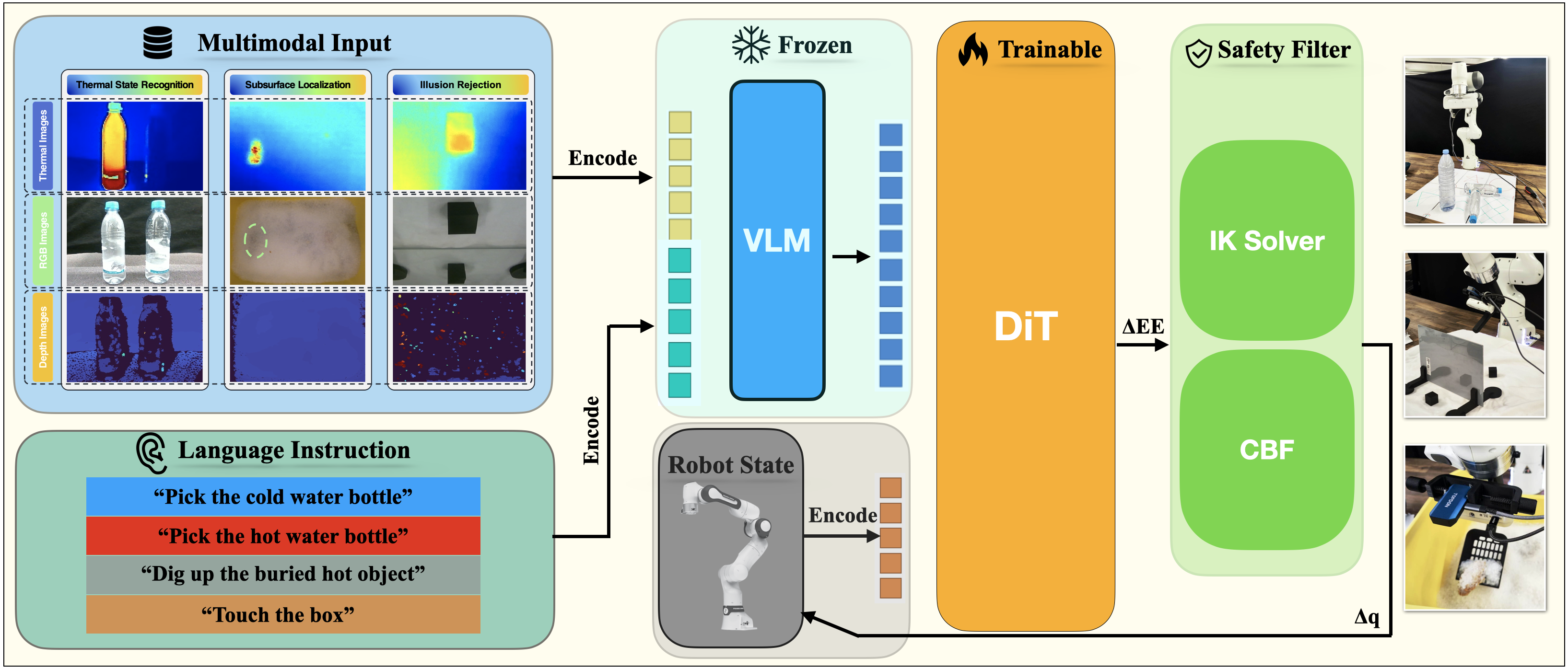}
     \caption{System Architecture of Safe-Night VLA. The pipeline contains synchronized RGB-D and LWIR inputs, frozen VLM feature extraction, the DiT action head, and the CBF-QP runtime safety layer used for hardware execution.
    }
    \label{fig:system_overview}
\end{figure}

The proposed Safe-Night VLA framework (\cref{fig:system_overview}) integrates thermal and depth perception into an existing RGB-pretrained VLA model. We instantiate the framework using the GR00T-N1.5-3B architecture \cite{nvidia2025gr00tn1openfoundation}. The core perception engine, EAGLE 2.5 \cite{chen_NeurIPS_2025_Eagle25}, fuses a SigLIP-2 400M vision encoder with a Qwen3-1.7B large language model \cite{yang2025qwen3technicalreport}. Visual tokens are projected into the LLM's embedding space via an MLP connector to form a joint visual-linguistic representation. The policy head is a 16-layer Diffusion Transformer (DiT) \cite{peebles2023scalable} formulated via Flow Matching. 
Conditioned on the multimodal features via cross-attention, the DiT predicts a 1-step action $\mathbf{a}$, consisting of 3D translation, 3D axis-angle rotation, and a binary gripper command.

Rather than designing a multimodal architecture from scratch, we adapt the foundation model to process non-RGB modalities via a simple parameter-efficient strategy. The VLM backbone, comprising the vision encoder and language model, is strictly frozen to preserve pre-trained semantic representations. Training is restricted exclusively to the action head components (the Vision-Language LayerNorm projector and the DiT weights). 
Empirical results indicate that the frozen vision encoder possesses sufficient shape and intensity bias to transfer effectively to the pseudo-color thermal and depth distributions without requiring structural modifications or Low-Rank Adaptation (LoRA). 

\subsection{Multimodal Input Processing}
The system processes three synchronized camera views: RGB, thermal, and depth. 
To leverage the frozen pre-trained vision encoder without architectural modifications, thermal and depth data are formatted as 3-channel pseudo-color images:
\begin{enumerate}
    \item {RGB ($I_{rgb}$):} Standard color imagery providing global visual context.
    \item {Thermal ($I_{therm}$):} The native pseudo-color output (e.g., Iron/Rainbow palette) from the thermal camera is cropped and aligned to the workspace. The temperature gradients are represented as 3-channel color intensities.
    \item {Depth ($I_{depth}$):} Raw depth maps are clipped to a 5-meter range and mapped to a Turbo colormap \cite{mikhailov2019turbo}, providing illumination-invariant geometric cues.
\end{enumerate}
These three views are passed to the vision encoder as independent image tokens, along with the text instruction. 
The thermal stream is used as a relative pseudo-color observation rather than as calibrated radiometric thermometry. 
Consequently, if the task-defining thermal contrast fully vanishes because the target and background reach thermal crossover, or because emissivity and ambient conditions eliminate the observable LWIR difference, passive thermal input alone cannot recover the missing physical signal.
For modality-ablation experiments, disabled modalities are not removed from the input structure; instead, they are replaced by zero-mask image tensors so that all model variants preserve the same token layout.

Moreover, we introduce an asymmetric data augmentation pipeline to reduce the policy's over-reliance on standard visual features. During training, severe photometric perturbations, including random brightness variations, color jitter, Gaussian noise, and random cropping, are applied exclusively to the RGB view ($I_{rgb}$).
The thermal and depth views receive only deterministic resizing. This procedure mathematically simulates unpredictable illumination conditions, naturally encouraging the model to attend to the domain-invariant thermal and geometric representations for task completion.

\subsection{Safety Guarantee}
Existing works integrating CBF with diffusion- or flow matching-based generative models typically apply safety filtering as a post-hoc projection of the planned Cartesian or joint-space trajectories onto a feasible safe set~\cite{dai_arxiv_2025_safeflowsaferobotmotion,yang_arxiv_2026_uniconflowunifiedconstrainedflowmatching}. 
However, while the VLA policy in our framework predicts a 7-dimensional action comprising a desired Cartesian end-effector (EE) delta pose $\boldsymbol{u}_\text{vla} \in \mathbb{R}^6$ and a binary gripper command $c_\text{grip}$, such end-to-end generative policies provide no inherent guarantees of geometric or kinematic safety during execution.

To ensure safe execution with CBFs, we decouple the gripper actuation from the spatial motion, where a joint-space runtime safety filter formulated as a strictly convex Quadratic Program (QP) is implemented. 
At each control step, given the measured joint configuration $\boldsymbol{q} \in \mathbb{R}^7$, the QP solves directly for the safe joint displacement $\Delta \boldsymbol{q}_\text{safe} \in \mathbb{R}^7$ to track the 6-DoF spatial intent $\boldsymbol{u}_\text{vla}$:
\begin{subequations}
\label{eq:cbf_qp}
\begin{align}
    \Delta \boldsymbol{q}_\text{safe} &= \operatorname*{argmin}_{\Delta \boldsymbol{q} \in \mathbb{R}^7} \, \lVert \boldsymbol{J}(\boldsymbol{q}) \Delta \boldsymbol{q} - \boldsymbol{u}_\text{vla} \rVert^2 + \lambda \lVert \Delta \boldsymbol{q} \rVert^2, \\
    \textrm{s.t.} \quad & \nabla h_{\text{col}}(\boldsymbol{q})^\top \Delta \boldsymbol{q} + \gamma h_{\text{col}}(\boldsymbol{q}) \geq 0, \\
    & \boldsymbol{q}_{\text{min}} \leq \boldsymbol{q} + \Delta \boldsymbol{q} \leq \boldsymbol{q}_{\text{max}}.
\end{align}
\end{subequations}

In this unified formulation, the current state $\boldsymbol{q}$ is treated as a known constant parameter. Consequently, the geometric Jacobian $\boldsymbol{J}(\boldsymbol{q}) \in \mathbb{R}^{6 \times 7}$ and the constraint gradient $\nabla h_{\text{col}}(\boldsymbol{q})$ are evaluated pointwise, rendering the objective quadratic and the constraints strictly linear with respect to the decision variable $\Delta \boldsymbol{q}$. 
The objective function seamlessly embeds a differential inverse kinematics solver, minimizing the tracking error between the physical robot's motion and the VLA's Cartesian intent $\boldsymbol{u}_\text{vla}$. 
The regularizer $\lambda > 0$ yields a damped least-squares formulation, which not only penalizes excessive joint velocities but also ensures the Hessian matrix remains strictly positive definite, guaranteeing numerical stability near singular configurations.
The continuously differentiable function $h_{\text{col}}(\boldsymbol{q}): \mathbb{R}^7 \rightarrow \mathbb{R}$ evaluates the minimum distance between the environment and a set of collision spheres attached to the manipulator links.

The safety guarantee is conditional on the modeled safe set, accurate robot state estimation, and correct geometric parameters for the workspace constraints. 
Because the safety filter is decoupled from the generative VLA policy, it may also repeatedly block or project similar infeasible actions. 
The CBF-QP layer should therefore be interpreted as a deterministic execution safeguard for modeled constraints.

\section{EXPERIMENTS}

\subsection{Experimental Setup}

\subsubsection{Dual-Arm Teleoperation Platform}
Our robotic platform is established with Franka Emika Panda 7-DOF manipulators. 
To gather expert demonstrations, we developed a master-slave dual-arm teleoperation system. The master arm's joint positions and gripper states are read at 500\,Hz, and converted into 1000\,Hz commands for the slave arm. 
The operator visually monitors the workspace to dynamically adjust movements, ensuring precise data collection.

\subsubsection{Sensor Setup and Input Processing}
The perception system consists of a side-mounted Intel RealSense RGB-D camera and a low-cost Topdon TC001 thermal camera. 
All three views are resized to $224 \times 224$ before being fed as separate image tokens to the VLM backbone.
Each collected data sample consists of a tuple $\bigl(\{I_\text{rgb}, I_\text{therm}, I_\text{depth}\}, \boldsymbol{q}, T_\text{prompt}, \boldsymbol{a}\bigr)$, where $T_\text{prompt}$ is the natural language instruction.

\subsubsection{Data Collection}
We collected a total of 600 expert demonstrations across the three scenarios, with each episode lasting approximately 200 state-action pairs. 
To ensure the model learns targeted semantic meanings, we utilized specific natural language instructions corresponding to the physical and thermodynamic goals of each task. \cref{tab:dataset_specs} details the task specifications, the exact language prompts used, and the data volume. 

\begin{table}[h]
\centering
\caption{Dataset Specifications}
\label{tab:dataset_specs}
\resizebox{\columnwidth}{!}{
\begin{tabular}{l p{5.2cm} c}
\toprule
\textbf{Task Name} & \textbf{Language Instructions (Prompts)} & \textbf{Episodes} \\
\midrule
\Romannum{1}. Temperature-conditioned manipulation & ``Pick the hot water bottle'' & 100 \\
\Romannum{1}. Temperature-conditioned manipulation & ``Pick the cold water bottle'' & 100 \\
\Romannum{2}. Subsurface target localization & ``Dig up the buried hot object'' & 200 \\
\Romannum{3}. Reflection disambiguation & ``Touch the box'' & 200 \\
\bottomrule
\end{tabular}
}
\end{table}

\subsubsection{Training Configuration}
We fine-tune the GR00T-N1.5-3B foundation model for 5,000 steps with a frozen VLM backbone. 
We strictly limit gradient updates to the Action Head components (the VLM projector and the Diffusion Transformer). No LoRA adapters are used.
To evaluate the contribution of each modality fairly, we train four separate model variants under the same architecture, frozen-backbone strategy, and optimization schedule: {RGB-Only}, {RGB-D}, {RGB-T}, and the Safe-Night VLA {RGB-T-D} model. Missing modalities are consistently replaced by zero-mask image tensors during both training and inference. Concretely, the RGB-Only model receives zero-masked thermal and depth inputs; RGB-D receives a zero-masked thermal input; RGB-T receives a zero-masked depth input; and the full model uses all three modalities. Thus, each modality configuration corresponds to an independently trained model rather than an inference-time ablation of a single shared checkpoint.

Training employs the AdamW optimizer ($\beta_1=0.95, \beta_2=0.999, \epsilon=10^{-8}$) with a weight decay of $10^{-4}$. We use a learning rate of $3 \times 10^{-5}$ with a cosine decay schedule and a 5\% linear warmup. The per-GPU batch size is set to 16 with 8 gradient accumulation steps, resulting in an effective batch size of 128. Training is conducted in bfloat16 mixed precision with TF32 enabled.

\subsubsection{Inference and Hardware Deployment}
During inference, the VLA's predicted Cartesian delta pose is directly processed by the unified CBF-QP solver to compute a safe joint trajectory. 
To evaluate robustness, we use the RGB input stream's brightness in real time to simulate dim/night environments without altering physical thermodynamics. 
We use the thermal camera's default pseudo-color mapping and do not perform extensive radiometric calibration; therefore, the reported dim/night results isolate the effect of degraded visible-light perception under otherwise unchanged thermal and geometric conditions.

\subsection{Scenario \Romannum{1}: Temperature-Conditioned Manipulation}
\begin{figure}[t]
    \centering
    \includegraphics[width=0.85\linewidth]{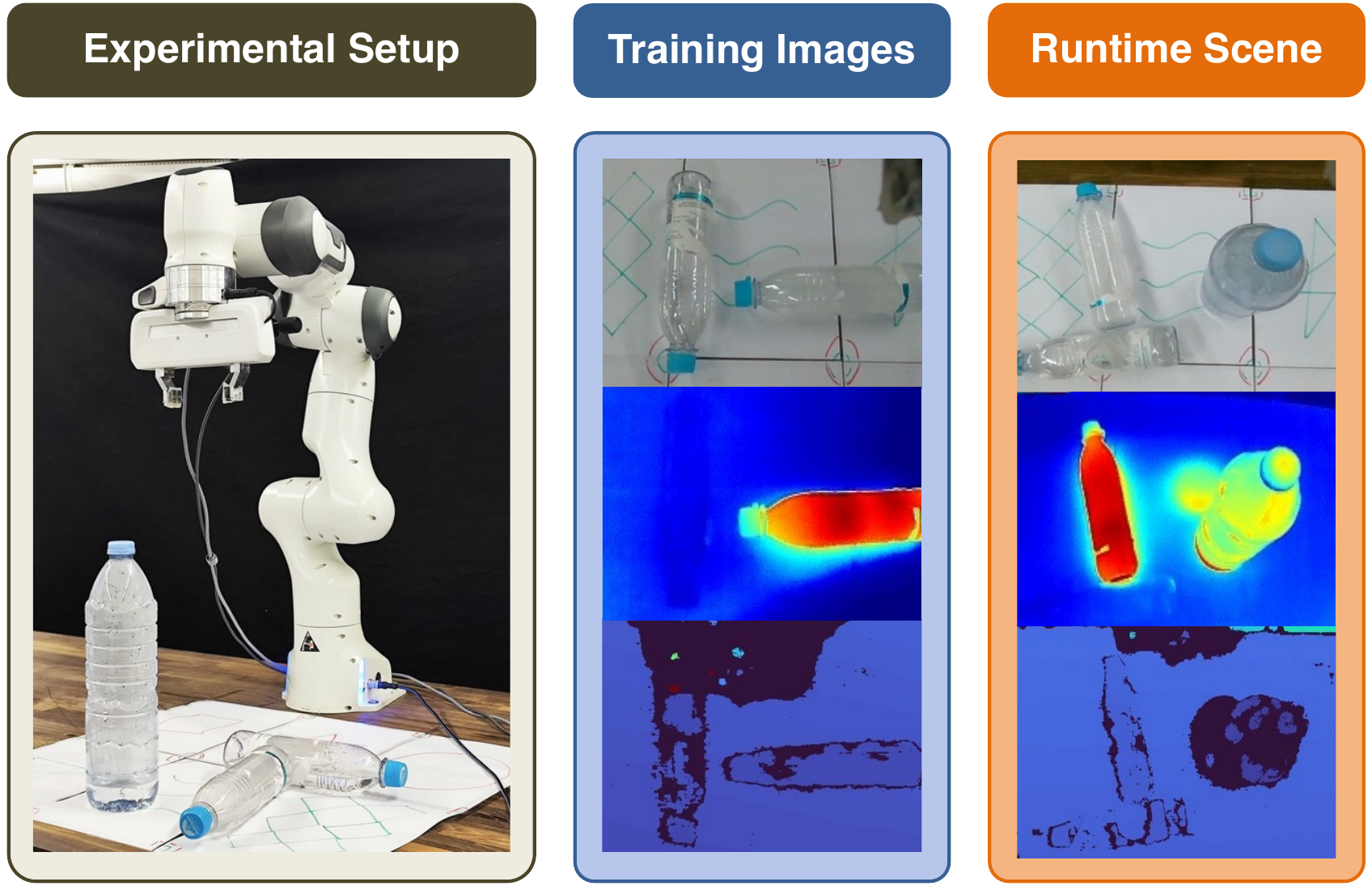}
    \caption{Temperature-Conditioned manipulation.}
    \label{fig:scen1}
\end{figure}

This scenario evaluates the robot's ability to ground temperature-based semantic adjectives into physical actions. During training, the workspace contains two physically identical bottles (either opaque or transparent); one is filled with hot water ($\sim 60^\circ$C) and the other with cold water ($\sim 20^\circ$C). The policy is conditioned on language instructions such as ``Pick the hot water bottle'' or ``Pick the cold water bottle,'' and outputs a 6-DoF EE delta pose alongside a binary gripper command. As illustrated in Fig.~\ref{fig:scen1}, at runtime we further introduce an additional upright heated object that is absent from the training demonstrations. This creates a distribution shift in which the policy must still select the language-specified target bottle rather than simply moving toward the most salient heat source in the scene. Meanwhile, the runtime safety filter constrains the executed motion to avoid unsafe behavior around this unseen heat-source obstacle.

As illustrated in Fig.~\ref{fig:concept_teaser}, standard RGB sensors capture only surface reflectance, making the hot and cold bottles visually identical. In contrast, thermal imaging captures emitted LWIR radiation, revealing a distinct high-intensity signature for the hot bottle. Providing this pseudo-color representation allows the frozen vision backbone to naturally associate semantic tokens (``hot''/``cold'') with the corresponding thermodynamic state, resolving the visual ambiguity.

\subsection{Scenario \Romannum{2}: Subsurface Localization}
\begin{figure}[t]
    \centering
    \includegraphics[width=0.85 \linewidth]{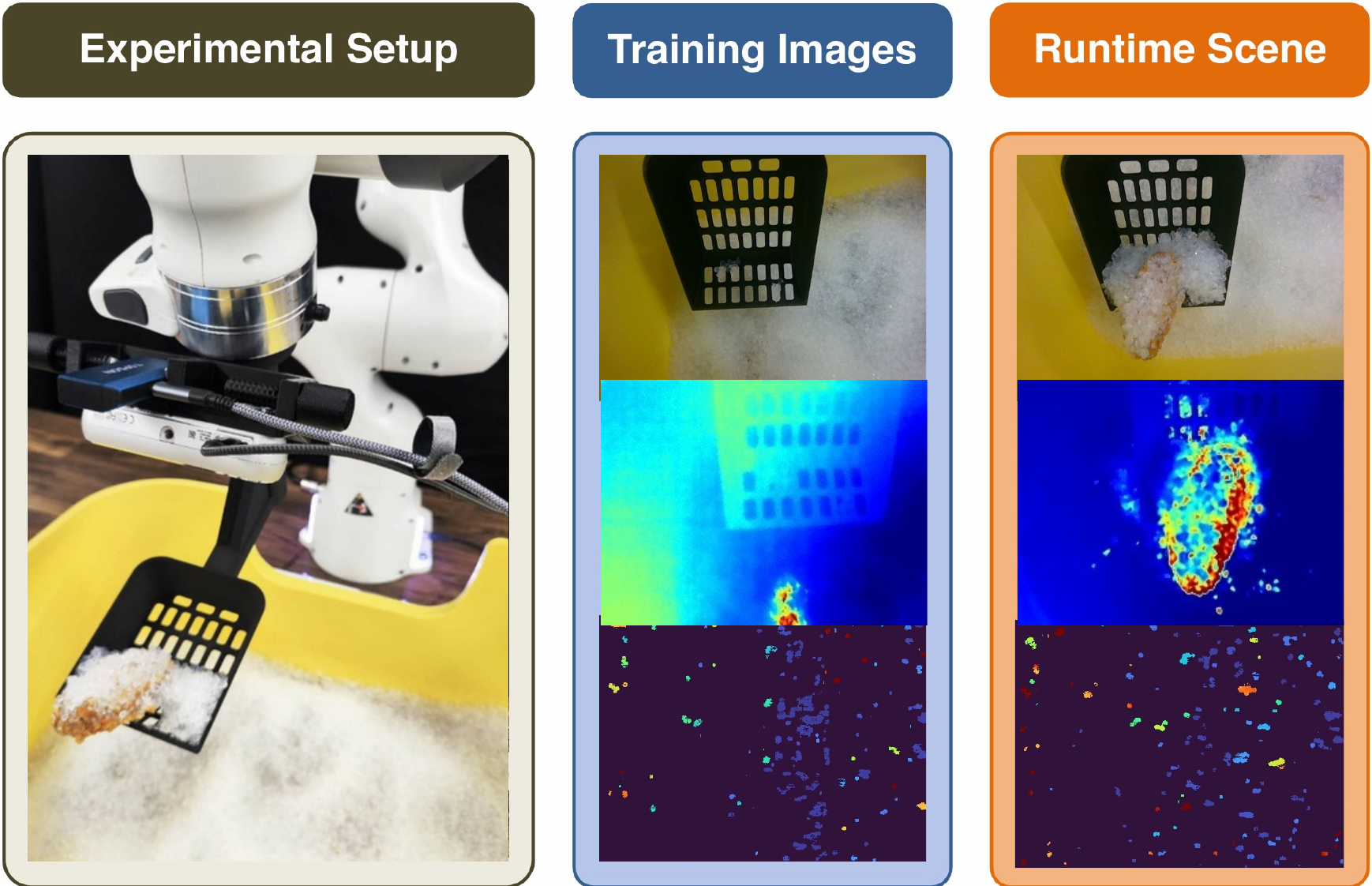}

    \caption{Subsurface Localization.}
    \label{fig:scen2}
\end{figure}

This scenario evaluates the system's ability to localize targets that are severely occluded by granular media. The robot, equipped with a specialized 3D-printed scoop EE, is tasked with localizing and unearthing a heated target (a chicken wing at $\sim 50^\circ$C) buried under $0.5$\,cm of crystal cat litter inside a box container. Prompted with the instruction ``Dig up the buried hot object,'' the policy must predict the subsurface target location and execute a plunging trajectory into the material. To strictly decouple the evaluation of multimodal localization capability from the mechanical dexterity required for granular scooping, we adopt a softened success criterion for this task: a trial is considered successful if the robot accurately locates and physically unearths the buried target (e.g., dislodging or lifting it to the surface), even if the object does not perfectly land inside the scoop.

During data collection, the language prompt specifies only the buried hot object rather than the surrounding box geometry. Nevertheless, the expert demonstrations implicitly encode the box as a geometric constraint, since the trajectories consistently avoid colliding with the litter-box boundary. At deployment time, the same physical structure remains present, and the runtime safety filter further enforces collision-free execution with respect to the box boundary.

As shown in Fig.~\ref{fig:concept_teaser}, the granular medium obscures the target in the visible spectrum. However, according to heat transfer principles \cite{kaviany2012principles}, thermal energy from the submerged source diffuses through the granular material, producing a detectable surface ``thermal bloom''. By capturing this emitted radiation via the LWIR sensor, the policy aligns the surface thermal pattern with the required subsurface localization.

\subsection{Scenario \Romannum{3}: Cross-Modal Disambiguation}

\begin{figure}[h!]
    \centering
    \includegraphics[width=0.85\linewidth]{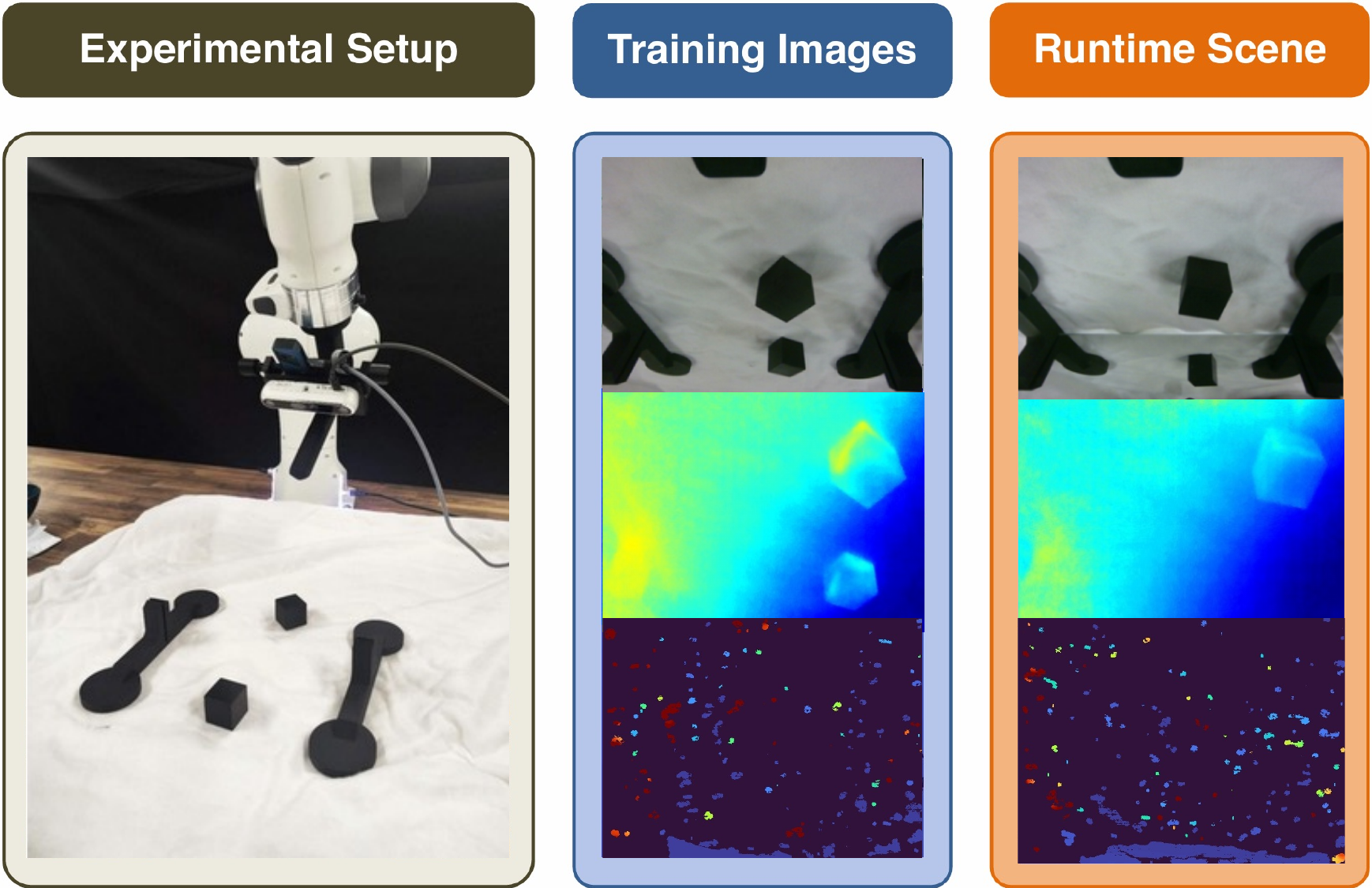}
    \caption{Cross-modal disambiguation under mirror-induced ambiguity.}
    \label{fig:scen3}
\end{figure}

This scenario investigates the system's ability to resolve cross-modal sensory conflict under OOD visual conditions. The robot, equipped with a stick EE, is prompted with a generic instruction: ``Touch the box.'' The training dataset contains no mirrors and consists only of two physical conditions: a ``Two Boxes'' setup, where the operator touches either real box with equal probability, and a ``Single Box'' setup with only one real object. Consequently, the policy is trained entirely on real object configurations and never observes mirror-induced false targets.

As illustrated in Fig.~\ref{fig:scen3} and in the right panel of Fig.~\ref{fig:concept_teaser}, the mirror produces a specular reflection in the visible spectrum, heavily resembling the ``Two Boxes'' training setup. However, standard glass and silver mirrors are largely opaque to LWIR radiation \cite{rogalski2000infrared, vollmer2017infrared}. Instead of reproducing the thermal emission, the mirror surface remains near room temperature, allowing the thermal modality to present a clear ``Single Box'' state and reject the reflected false target.

This scenario also exposes a second execution-level failure mode unrelated to semantic target selection alone. In our setup, we impose a modeled wall constraint located 5\,cm behind the initial EE position along the positive $x$ direction. The demonstration dataset contains essentially no backward EE motion in this direction. Under perceptual ambiguity or sampling noise, however, the VLA can occasionally propose extrapolative backward actions that move the EE toward this unseen wall. The runtime safety filter prevents such unsafe motion by enforcing the wall constraint at execution time, ensuring that noisy OOD commands do not produce a hardware collision.

The evaluation is formulated as a 1-step open-loop action prediction rather than a closed-loop physical touching sequence. In the expert demonstrations, targeting the far object is strictly mapped to a fixed 20\,cm forward movement, while targeting the near object is mapped to a fixed 20\,cm downward movement. 
Consequently, performance is evaluated based on the predicted trajectory rather than physical contact, featuring two distinct failure modes: {(i) Directional Errors}, where the robot fails to output the designated 20\,cm directional movement corresponding to the real physical target, and {(ii) Deception Rate}, where the policy predicts a trajectory oriented toward the virtual reflection. In addition, any action proposal that would violate the modeled backward wall constraint is intercepted by the safety filter and recorded as a failed trial rather than a successful recovery.
Notably, no manual action selection, online correction, or human labeling of trajectories is used during this evaluation. The first predicted EE displacement is used as the diagnostic signal: the forward 20\,cm template indicates attraction to the far physical target, the downward 20\,cm template indicates attraction to the near physical target, and a reflection-oriented template is counted as deception. We deliberately use this open-loop protocol to isolate mirror-induced policy deception; a closed-loop path executor could mask an initially wrong action through downstream recovery and is therefore left for system-level follow-up work.

\section{RESULTS}

\begin{table*}[t]
\centering
\caption{Joint ablation of input modalities and runtime safety filtering under normal and dim/night illumination.}
\label{tab:results}
\resizebox{\textwidth}{!}{
\begin{tabular}{lcccccccccccccccc}
\toprule
& \multicolumn{8}{c}{\textbf{Normal Light}} & \multicolumn{8}{c}{\textbf{Dim/Night Light}} \\
\cmidrule(lr){2-9} \cmidrule(lr){10-17}
\textbf{Task \& Metric}
& \multicolumn{4}{c}{\textbf{w/o Safety Filter}}
& \multicolumn{4}{c}{\textbf{w/ Safety Filter}}
& \multicolumn{4}{c}{\textbf{w/o Safety Filter}}
& \multicolumn{4}{c}{\textbf{w/ Safety Filter}} \\
\cmidrule(lr){2-5} \cmidrule(lr){6-9} \cmidrule(lr){10-13} \cmidrule(lr){14-17}
& \textbf{RGB-Only} & \textbf{RGB-D} & \textbf{RGB-T} & \textbf{Ours}
& \textbf{RGB-Only} & \textbf{RGB-D} & \textbf{RGB-T} & \textbf{Ours}
& \textbf{RGB-Only} & \textbf{RGB-D} & \textbf{RGB-T} & \textbf{Ours}
& \textbf{RGB-Only} & \textbf{RGB-D} & \textbf{RGB-T} & \textbf{Ours} \\
\midrule

\multicolumn{17}{l}{\textit{Success Rate (SR) $\uparrow$ (N = 50 trials)}} \\
\addlinespace[2pt]

1. Hot/Cold Bottle
& 32\% & 24\% & \textbf{78\%} & 72\%
& 42\% & 38\% & \textbf{86\%} & 82\%
& 0\% & 10\% & 10\% & \textbf{56\%}
& 0\% & 12\% & 22\% & \textbf{64\%} \\

2. Buried Object
& 14\% & 24\% & 60\% & \textbf{74\%}
& 16\% & 24\% & 66\% & \textbf{78\%}
& 0\% & 0\% & 42\% & \textbf{68\%}
& 0\% & 2\% & 48\% & \textbf{72\%} \\

\addlinespace[3pt]
\multicolumn{17}{l}{\textit{Success Counts $\uparrow$ (N = 20 trials per sub-case)}} \\
\addlinespace[2pt]

3a. Single Box Success
& \textbf{13/20} & 8/20 & 10/20 & 9/20
& 19/20 & 17/20 & \textbf{20/20} & 17/20
& 0/20 & 0/20 & 4/20 & \textbf{12/20}
& 9/20 & 9/20 & 17/20 & \textbf{18/20} \\

3b. Mirror Rejection Success
& 2/20 & 3/20 & \textbf{12/20} & 4/20
& 12/20 & 11/20 & \textbf{19/20} & 15/20
& 3/20 & 0/20 & \textbf{7/20} & 5/20
& 13/20 & 11/20 & 15/20 & \textbf{17/20} \\

\bottomrule
\end{tabular}
}
\end{table*}

\subsection{Quantitative Performance Comparison}
We evaluated four separately trained modality configurations across the three scenarios under two illumination conditions, \textit{Normal Light} and \textit{Dim/Night Light}, and under two execution settings: \textit{w/o Safety Filter} and \textit{w/ Safety Filter}. The tested variants include \textit{RGB-Only}, \textit{RGB-D}, \textit{RGB-T}, and our \textit{Safe-Night VLA} model. For each variant, unavailable modalities are consistently replaced with zero-mask image inputs during both training and inference, rather than being removed only at test time.

As summarized in Table~\ref{tab:results}, the ablation reveals a clear separation between semantic perception and execution safety. Across Scenarios \Romannum{1} and \Romannum{2}, variants equipped with thermal input substantially outperform RGB-Only and RGB-D baselines, confirming that thermal sensing provides the dominant task-relevant signal for thermodynamic discrimination and subsurface target localization. 
In Scenario \Romannum{1} ({Hot/Cold Bottle}), RGB-T achieves the strongest normal-light performance without the safety filter (78\%), while enabling the filter further improves both thermal-equipped variants (RGB-T: 86\%, Ours: 82\%), indicating that a non-trivial fraction of failures arises from execution-level geometric conflicts near the tabletop and the nearby cylindrical heat-source obstacle rather than target misidentification alone. Under dim/night conditions, the Safe-Night VLA becomes the most robust configuration (64\%), showing that thermal remains the primary semantic cue while depth contributes complementary geometric stabilization once RGB structure is degraded.

In Scenario \Romannum{2}, the same modality trend holds: thermal-equipped variants remain substantially stronger than RGB-Only and RGB-D under both illumination conditions, and the full model achieves the highest overall success (78\% in normal light and 72\% in dim/night light with the safety filter). Compared with Scenario \Romannum{1}, however, the relative gain from the safety filter is more modest. 
This suggests that the dominant challenge in this task is still subsurface target localization rather than severe OOD motion proposals. 
Notably, in many failed trials near the litter-box boundary, being blocked by the CBF filter and physically colliding with the box led to the same outcome.

Scenario \Romannum{3} exhibits the strongest dependence on runtime safety. Here we report {Single Box Success} and {Mirror Rejection Success}. Under normal lighting, thermal input is already highly beneficial for rejecting the reflected false target: without the safety filter, RGB-T reaches 12/20 mirror-rejection successes, compared with only 2/20 for RGB-Only and 3/20 for RGB-D. However, enabling the safety filter sharply boosts success for all variants, especially the thermal-equipped ones (RGB-T: 12/20 $\rightarrow$ 19/20; Ours: 4/20 $\rightarrow$ 15/20). The effect becomes even more pronounced under dim/night conditions, where the full model improves from 12/20 to 18/20 on {Single Box Success} and from 5/20 to 17/20 on {Mirror Rejection Success}. This indicates that once RGB input is heavily attenuated, the policy is more likely to produce unstable, extrapolative EE commands, including backward motions that were absent from the demonstration set and therefore directly conflict with the modeled wall constraint. 
In this case, the safety filter is not merely a collision-prevention layer, but it is a critical execution safeguard against geometrically unsafe OOD actions.

Overall, the results show that thermal is the most informative modality for resolving hidden physical states and optical ambiguity in our setup, while the safety filter and depth input primarily improve the probability that semantically correct decisions can be executed robustly under degraded illumination and modeled geometric constraints.

\subsection{Result Analysis and Failure Modes}

Beyond the aggregate metrics, the failure patterns reveal a consistent division of labor across modalities and safety layers: thermal input primarily resolves hidden physical states and cross-modal ambiguity, whereas depth and the runtime safety filter mainly improve execution robustness when illumination is degraded or the policy encounters modeled geometric constraints.

\subsubsection{Limitations of RGB-Only and RGB-D Variants}
The non-thermal baselines exhibit two systematic vulnerabilities.

\begin{itemize}
    \item {Insufficient physical-state observability (Scenarios \Romannum{1} \& \Romannum{2}):} In the {Hot/Cold Bottle} task, RGB-Only and RGB-D remain weak even under normal light, indicating that geometric and appearance cues alone are insufficient to reliably ground the semantic distinction between thermodynamically different but visually similar objects. This limitation is even more severe in the {Buried Object} task, where both variants remain weak in normal light and nearly collapse under dim/night conditions.
    
    \item {Sensitivity to high-uncertainty visual inputs (Scenario \Romannum{3}):} In the mirror setup, RGB-Only and RGB-D are frequently misled by reflected false targets under normal lighting, and degrade further under dim/night conditions. Once RGB information is strongly attenuated, these variants often exhibit poorly conditioned open-loop behavior, including spatially unstable action proposals and, in some cases, backward EE increments that are outside the support of the training demonstrations. 
\end{itemize}

\subsubsection{Thermal Semantics, Runtime Safety, and Low-Light Robustness}
Adding thermal input substantially changes the model behavior, but safe execution still depends on the interaction between perception and runtime constraints.

\begin{itemize}
    \item {Thermal as the dominant semantic modality (Scenarios \Romannum{1} \& \Romannum{2}):} The RGB-T variant strongly improves success in both {Hot/Cold Bottle} and {Buried Object}, showing that thermal input provides the decisive signal for hidden-state reasoning. In Scenario \Romannum{1}, thermal cues are sufficient to resolve the semantic ambiguity between hot and cold targets; in Scenario \Romannum{2}, the policy can exploit the surface thermal bloom induced by conductive heat transfer through the granular medium.
    
    \item {Depth and safety mainly stabilize execution under degraded illumination:} Although RGB-T is strongest in several normal-light settings, the Safe-Night VLA model becomes the most robust configuration under dim/night conditions. This pattern suggests that depth does not act as the primary semantic discriminator, but instead provides complementary geometric stabilization once RGB cues become unreliable. The runtime safety filter plays a similar role at the control level, suppressing unstable OOD motions that become more frequent when perception is uncertain.
    
    \item {Mirror disambiguation is mainly driven by thermal input, but execution safety remains essential (Scenario \Romannum{3}):} In normal light, RGB-T already achieves strong mirror rejection, indicating that thermal input is the principal cue for suppressing reflected false targets, since standard mirrors do not reproduce the object's LWIR signature. However, without the safety filter, this semantic advantage does not always translate into task success, because the policy can still generate geometrically unsafe motions under the added wall constraint. Under dim/night conditions, this issue becomes more severe: the degradation of RGB structure increases action uncertainty, and the safety filter becomes critical for converting semantically correct target selection into successful physical execution.
    
    \item {Residual failures arise from recovery limitations and weak negative thermal contrast:} Not all failures are caused by semantic misclassification. In the {Hot/Cold Bottle} and {Buried Object} tasks, a common failure mode occurs when the EE approaches the boundary too closely. In these cases, the CBF layer correctly blocks the unsafe motion, but the policy often continues proposing similar downward actions instead of lifting, re-centering, and retrying. As a result, being safely stopped by the filter and physically colliding with the boundary can still produce the same task-level failure. Under dim/night conditions, the challenge is further amplified for the {cold} bottle, whose temperature may lie close to ambient background levels; once RGB structure is strongly attenuated, the weak negative thermal contrast makes localization less reliable even when thermal input is present. 
    A closed-loop extension could feed the CBF intervention flag back into the policy, trigger a traceback primitive along recent safe states, or invoke retry demonstrations when repeated infeasible commands are detected. 
    For weak thermal contrast, ambient-reference normalization, local contrast enhancement, background subtraction, radiometric calibration, and temporal filtering are natural preprocessing extensions.
\end{itemize}

\subsection{Exploratory Mechanism Analysis: Attention Ablation}

To explore how the policy utilizes thermal cues rather than relying on dataset-induced spatial biases, we conduct a preliminary attention ablation study on Scenario \Romannum{1} across 10 execution episodes (prompt: ``pick hot water''). We contrast the internal representations generated with valid thermal inputs against those with explicitly masked thermal channels. 
Specifically, we extract spatial attention maps by combining the frozen vision encoder's Grad-CAM saliency with the DiT action head's token-level cross-attention. 
We evaluate two metrics: Attention Concentration ($E$), measured by the normalized entropy of the spatial distribution (where lower values indicate focused localization); and Semantic Alignment ($r$), the Pearson correlation between the spatial map and ground-truth thermal pixel intensities.

\begin{figure}[h!]
    \centering
    \includegraphics[width=0.9\linewidth]{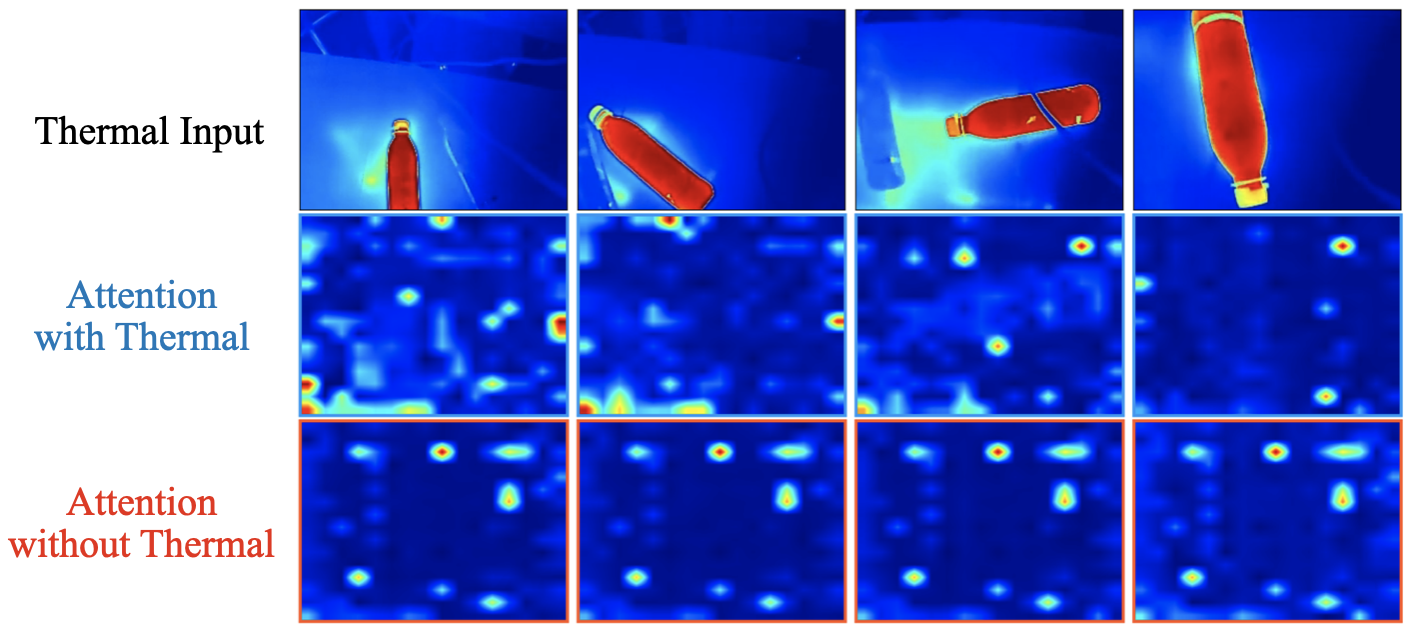} 
    \caption{Attention ablation in Scenario \Romannum{1}.}
    \label{fig:attention}
\end{figure}

The comparative metrics across the sampled episodes suggest specific cross-modal mapping behaviors. {Concentration:} With thermal input, the normalized entropy $E$ decreases significantly ({0.052} vs. 0.228 without thermal data), tightly bounding the target object (Fig. \ref{fig:attention}). Without it, the distribution scatters. {Alignment \& Target Mass:} The Pearson correlation $r$ shifts from a negative baseline ($-0.064$) to positive ({$+0.081$}) with thermal input, and the proportion of attention mass on the hot object increases from 16.8\% to {53.5\%}. Crucially, these findings do not imply that the VLA performs complex thermodynamic reasoning. Rather, they suggest that the frozen RGB vision encoder successfully transfers its pre-trained visual saliency bias to the pseudo-color thermal domain. The action head learns to correlate the semantic token ``hot'' with the high-intensity pixel regions generated by thermal emissions, effectively locating the heat source.

\section{CONCLUSION}
In this work, we presented {Safe-Night VLA}, a multimodal manipulation framework that addresses the fundamental limitations of RGB-only policies by integrating thermal perception with safety guarantees. 
Rather than treating thermal imaging merely as a low-light heuristic, we demonstrated its capability to ground unobservable physical states, enabling the system to distinguish thermodynamic properties, localize subsurface targets, and reject optical illusions.

While our empirical study demonstrates the critical value of thermal perception, we acknowledge several limitations:
First, the evaluated tasks serve as targeted diagnostic scenarios designed specifically to expose the blind spots of RGB-only policies, rather than an open-ended general manipulation benchmark. Similarly, the natural language instructions primarily evaluate physically grounded task conditioning rather than broad language generalization. 
Second, our empirical evaluations are grounded in a pretrained model. We have yet to comprehensively benchmark the multimodal transferability and performance gains when scaling up to larger foundation models or exploring diverse VLA architectures. 
Last, the implemented CBF-QP acts as a runtime geometric safeguard. Its efficacy is contingent upon accurately modeled workspace boundaries and precise state estimation, and it currently does not account for unmodeled dynamic obstacles. 

Future work will focus on integrating real-time thermal point clouds to upgrade our static geometric safeguards into dynamic, temperature-aware obstacle avoidance. Furthermore, scaling this multimodal framework to SOTA, large-scale foundation models will be a key direction to unlock more robust manipulation in unstructured environments. We will also investigate recovery-aware closed-loop policies, CBF-intervention feedback, radiometric thermal calibration, and prompt-paraphrase evaluation for broader thermal-language generalization.

\section*{ACKNOWLEDGMENT}
This work was partially supported by the Huawei-TUM joint laboratory.

\bibliographystyle{IEEEtran}
\bibliography{ref_NoURL}

\end{document}